\documentclass[10pt, a4paper]{article}

\usepackage[final]{lrec2026} 
\usepackage{booktabs}
\usepackage{microtype}
\usepackage{expex}
\usepackage{amsmath}
\usepackage{CJKutf8}

\title{\textsc{CantoNLU}: A benchmark for Cantonese natural language understanding}

\name{
\begin{tabular}{@{}c@{}}
Junghyun Min\textsuperscript{$\dagger$} \quad
York Hay Ng\textsuperscript{$\ddagger$} \quad
Sophia Chan\textsuperscript{$\S$}\\
Helena Shunhua Zhao\textsuperscript{$\ddagger$} \quad
En-Shiun Annie Lee\textsuperscript{$\ddagger \&$}
\end{tabular}}

\address{\\
\textsuperscript{$\dagger$}Georgetown University \quad\quad 
\textsuperscript{$\ddagger$}University of Toronto\\
\textsuperscript{$\&$}Ontario Tech University \quad\quad
\textsuperscript{$\S$}Independent Researcher\\
jm3743@georgetown.edu }

\abstract{
Cantonese, although spoken by millions, remains under-resourced due to policy and diglossia. To address this scarcity of evaluation frameworks for Cantonese, we introduce \textsc{\textbf{CantoNLU}}, a benchmark for Cantonese natural language understanding (NLU).
This novel benchmark spans seven tasks covering syntax and semantics, including word sense disambiguation, linguistic acceptability judgment, language detection, natural language inference, sentiment analysis, part-of-speech tagging, and dependency parsing. In addition to the benchmark, we provide model baseline performance across a set of models: a Mandarin model without Cantonese training, two Cantonese-adapted models obtained by continual pre-training a Mandarin model on Cantonese text, and a monolingual Cantonese model trained from scratch. Results show that Cantonese-adapted models perform best overall, while monolingual models perform better on syntactic tasks. Mandarin models remain competitive in certain settings, indicating that direct transfer may be sufficient when Cantonese domain data is scarce. We release all datasets, code, and model weights to facilitate future research in Cantonese NLP.
 \\ \newline \Keywords{Cantonese, natural language understanding, transfer learning, benchmark} }

\begin{document}

\maketitleabstract

\section{Introduction}
\label{sec:intro}

Mandarin Chinese is considered a high-resource language, abundant with pre-trained language model (PLM) support \citep{devlin-etal-2019-bert, liu-etal-2020-multilingual-denoising, conneau-etal-2020-unsupervised}, corpora, and evaluation benchmarks \citep{xu-etal-2020-clue, xiang-etal-2021-climp}, and most recently commercial large language models \citep{bai2023qwen, liu2024deepseek}.
However, the same cannot be said about other variants of the Sinitic language family, including Cantonese. Mandarin is the official state language and the prestige language in media, business, and academia. Owing to this status, Cantonese, along with other Sinitic languages, remains a primarily vernacular language without written standardization \citep{snow2004cantonese, Li_2006-lingua-franca}.

Such lack of written standardization, prestige, and official status results in a shortage of language resources in Cantonese. It is  frequently described as a low-resource language \citep[e.g. ][]{liu-2022-low, xiang-etal-2022-cantonese, jiang2025developing} despite having millions of speakers \citep{ethnologue2023}. As a result, Cantonese language processing systems often rely on datasets, models, and corpora adapted from Mandarin \cite{xiang-etal-2024-cantonese}, despite a lack of mutual intelligibility between the two languages \citep{norman1988chinese, tang2007mutual}.

\begin{figure}[t]
\centering
\includegraphics[width=\columnwidth]{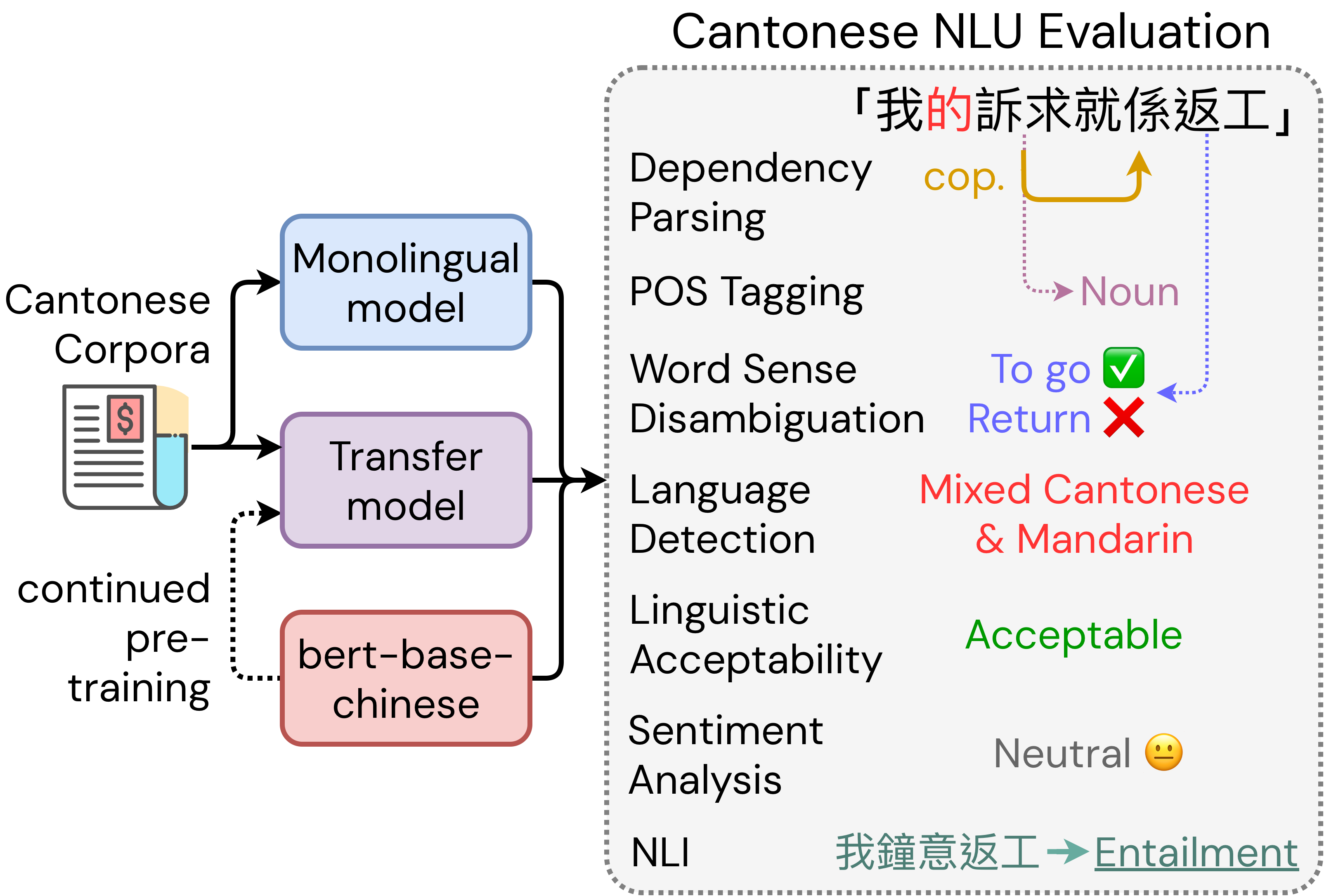}
\caption{An overview of the tasks in CantoNLU, and our framework for investigating when and how cross-lingual transfer learning from Mandarin is effective for natural language understanding in Cantonese.}
\label{fig:method}
\end{figure}

\begin{table*}[ht]
    \centering
    \begin{tabular}{llrl}
    \toprule
        Task & Requires & Size & Source\\
        \midrule
         \multicolumn{3}{l}{\textbf{Sentence-level tasks}}  \\
         Acceptability & L, F, M & 1.6k & MT error dataset \citep{liu2025siniticmterror} adapted \\
         Lang Detection & L, F & 47k & Parallel corpus \citep{dai-etal-2025-next} aligned and purturbed \\
         NLI & M & 570k & Machine-translated English NLI \citep{cheng2025yue-all-nli}\\
         Sentiment & M & 12k & Hong Kong restaurant reviews \citep{ZHANG2011OpenRice} \\
         \midrule
         \multicolumn{3}{l}{\textbf{Word-level tasks}}  \\
         WSD & L, M & 109 & Manual compilation\\
         POS, Dep parsing & F & 14k & Universal Dependencies dataset \citet{wong-etal-2017-quantitative}\\
         \bottomrule
    \end{tabular}
    \caption{An overview of the 7 NLU tasks that make up \textsc{\textbf{CantoNLU}}, what the task requires, their size and source. L, F, M stand for lexicon, form (syntax), and meaning (semantics), respectively.}
    \label{tab:stats}
\end{table*}

Cross-lingual transfer from Mandarin for Cantonese language processing has proven effective, with empirical success across a range of tasks, including language modeling and reading comprehension \citep{jiang2025developing}, translation \citep{liu-2022-low, suen-etal-2024-leveraging}, and speech recognition \citep{li2019cantonese, luo2021crosslanguage}. However, despite empirical success, there remains a gap in formal investigations into the best practices for Cantonese natural language understanding (NLU). This is attributable to a lack of a centralized evaluation framework for Cantonese language processing or understanding.

In this paper, we introduce \textsc{\textbf{CantoNLU}}, a GLUE-like \citep[General Language Understanding Evaluation; ][]{wang-etal-2018-glue} natural language understanding benchmark in Cantonese.
\textsc{\textbf{CantoNLU}} provides an in-depth evaluation of syntax, lexicon and semantic understanding, comprising 7 tasks: word sense disambiguation (WSD), linguistic acceptability judgment (LAJ), language detection (LD), natural language inference (NLI), sentiment analysis (SA), part-of-speech tagging (POS), and dependency parsing (DEPS).
In particular, the WSD dataset is entirely novel, providing the first resource for sense-level lexical understanding in Cantonese.
LAJ and LD represent novel adaptations of existing datasets, while NLI, SA, POS, and DEPS datasets are direct adoptions of existing datasets \citep[][respectively]{cheng2025yue-all-nli, ZHANG2011OpenRice, wong-etal-2017-quantitative}.
We describe each task and underlying datasets in Section \ref{sec:benchmark}, whose overview is outlined in Table \ref{tab:stats}.

Using \textsc{CantoNLU}, we evaluate three types of models -- a Mandarin model\footnote{\texttt{bert-base-chinese}} without explicit training on Cantonese, Cantonese-adapted transfer models with additional Cantonese training on a Mandarin model, and a monolingual Cantonese model, as outlined in Section \ref{sec:exp}.
In Section \ref{sec:results}, we discuss how monolingual Cantonese, Cantonese-adapted, and Mandarin models compare across various aspects in Cantonese NLU.

In spite of limited training data in Cantonese, we demonstrate that our monolingual Cantonese model excels in syntactic tasks such as POS and DEPS. 
On the other hand, Cantonese-adapted models excel in semantic tasks such as NLI, LD, WSD, and SA. 
Simultaneously, mandarin models offer a competitive alternative to additional or monolingual training on Cantonese data, excelling in NLI and LAJ.

Based on our results, we recommend monolingual Cantonese models for syntactic tasks, and Cantonese-adapted models for semantic tasks.
In domains where Cantonese corpora are scarce, Mandarin models without Cantonese adaptation may be sufficient. 
In addition to the benchmark and analysis, we publicly release our code and model weights at \url{github.com/aatlantise/sinitic-nlu}.

\section{Background}
\label{sec:background}

Cantonese, also known as Yue \citep{ethnologue2023}, is the second most widely used Sinitic language after Mandarin, spoken by an estimated 85 million people worldwide. 
However, it remains primarily a spoken language \citep{norman1988chinese}, with most speakers defaulting to written Standard Chinese \citep{xiang-etal-2024-cantonese}. 
This diglossic situation and its short history as a written language \citep{snow2004cantonese} has contributed to the scarcity of high-quality textual resources for Cantonese NLP, in stark contrast to Mandarin Chinese, a related language rich in resource across corpora, models \citep{devlin-etal-2019-bert, liu-etal-2020-multilingual-denoising, conneau-etal-2020-unsupervised, bai2023qwen, liu2024deepseek}, and evaluation resources \citep[e.g. ][]{xu-etal-2020-clue, xiang-etal-2021-climp}.
Thus, as highlighted in Section \ref{sec:intro}, prior work have used varying degrees of transfer from models with Mandarin knowledge to perform downstream tasks in Cantonese \citep{li2019cantonese, cheng2024bertbasecantonese, jiang2025developing}.

\paragraph{Cross-lingual transfer.}
Transfer learning is a common strategy in deep learning (DL) to address data sparsity, where features or signal learned from a resource-rich domain is used to augment a low-resource domain, task, or dataset \citep{pmlr-v27-bengio12a}.
In the context of NLP, a common application of transfer learning is in cross-lingual transfer for low-resource languages \cite{ruder-etal-2019-transfer, conneau-etal-2020-unsupervised}. 
Two types of cross-lingual transfer exists: model transfer and data transfer, as described in \citet{garcia-ferrero-etal-2022-model}.
Data transfer involves translating a dataset or corpus from a high-resource language to a lesser-resourced target language, as performed in \texttt{yue-all-nli} \citep{cheng2025yue-all-nli}, the MMLU \citep{hendrycks2021measuring} portion of the HKCanto-Eval benchmark \citep{cheng-etal-2025-hkcanto}, and Yue benchmarks from \citet{jiang-etal-2025-well}.

On the other hand, model transfer involves adapting models trained on a high-resource source language to a lesser-resourced target language, such as Mandarin to Cantonese and Danish to Faroese \citep{snaebjarnarson-etal-2023-transfer}.
Model transfer relies on the lexical or typological similarity between the two languages, thereby enabling the transfer of linguistic knowledge \cite{khan-etal-2025-uriel}.
While some implementations of cross-lingual transfer explicitly designate a source language for cross-lingual transfer \citep{snaebjarnarson-etal-2023-transfer, thangaraj2024crosslingualtransfermultilingualmodels}, others rely on multilingual models.
Rather than transferring from a specific source language, such models are thought to capture general cross-linguistic patterns that extend beyond typological or lexical similarity, allowing them to perform reasonably well even on unseen or low-resource languages \citep{conneau-etal-2020-unsupervised, scivetti-etal-2025-multilingual}.
Hybrid approaches combine both paradigms: they begin with a multilingual model, then continue pre-training or fine-tuning on a specific target language before applying the resulting model to downstream tasks \citep{protasov-etal-2024-super, manafi-krishnaswamy-2024-cross}.

Due to the richness of Mandarin language resources and the strong performance of Mandarin PLMs and LLMs, most cross-lingual transfer to Cantonese use Mandarin as the source language \citep{li2019cantonese, liu-2022-low, luo2021crosslanguage, suen-etal-2024-leveraging, cheng2024bertbasecantonese}, with exceptions using a multilingual model \citep[e.g.][]{jiang2025developing}.

However, there is emerging work suggesting limitations in models' ability to capture Cantonese idiosyncrasies.
Factors previously identified include substantial dissimilarities in lexicon, syntax and writing systems \cite{suen-etal-2024-leveraging}, along with the prevalence of colloquial phrases and code-switching in more recent Cantonese corpora \cite{jiang-etal-2025-well}.
These issues hinder the ability of Mandarin-trained models in Cantonese \cite{xiang-etal-2024-cantonese} due to over-reliance on Mandarin linguistic knowledge \cite{cheng-etal-2025-hkcanto}.

\paragraph{Cantonese grammar.}
Cantonese diverges from Mandarin in word order, particle and grammatical word inventory, and morphology, as highlighted in theoretical linguistics work \citep{matthews2006serial, yap2011asymmetry, matthews2013cantonese} as well as prior work in Cantonese-Mandarin machine translation and corpus linguistics \citep{zhang-1998-dialect, lam-2020-forms, liu2025siniticmterror}.
For example, in double object constructions, Mandarin takes the direct-indirect order as seen in (1), while Cantonese takes the indirect-direct order as seen in (2) \citep{matthews2013cantonese}.

\lingset{aboveglftskip=0pt}
\lingset{belowexskip=0.5ex}

\ex
\begingl
\gla gei3 ni3 qian2 //
\glb give you money //
\glft `(I) give you money' //
\endgl
\trailingcitation{(Mandarin)}
\xe

\ex
\begingl
\gla bei2 cin2 nei5 //
\glb give money you //
\glft `(I) give you money' //
\endgl
\trailingcitation{(Cantonese)}
\xe

Cantonese also features a substantially larger and more diverse inventory of particles and aspect markers than Mandarin, enabling speakers to encode subtle distinctions of tense, stance, and speaker attitude 
\citep{yap2011asymmetry, matthews2013cantonese}.
Cantonese morphology is more flexible, with more frequent verb serialization \citep{matthews2006serial} and reduplication \citep{LEE_2020_reduplication, lam-2020-forms} than in Mandarin.
This allows sequences such as (3), sourced from \citet{omelia1965first}, where three verbs combine to describe one scene and (4), where reduplicating the classifier (counting noun) yields an \textit{every} quantifier.
Given such unique properties of Cantonese grammar, availability of high-quality Cantonese data is critical to performing well on Cantonese linguistic tasks.

\ex
\begingl
\gla keoi5 jap6 heoi3 co5 //
\glb 3SG enter come sit //
\glft `He went in and sat down' //
\endgl
\xe

\ex
\begingl
\gla zek3 zek3 gau2 //
\glb CL CL dog //
\glft `every dog' //
\endgl
\xe

\section{Related Work}
\label{sec:related-work}

\paragraph{Cantonese language resources.}
Although widely considered low-resource, recent efforts have begun to address scarcity in Cantonese language resources.
Machine-translations of non-Cantonese resources may offer potential utility as data transfer, as provided by \citet{cheng2025yue-all-nli} for Cantonese natural language inference (NLI) and \citet{cheng-etal-2025-hkcanto, jiang-etal-2025-well} for LLM knowledge evaluation.
Foundational tools such as PyCantonese \citep{lee-etal-2022-pycantonese} provide \texttt{NLTK}-like \citep{bird-loper-2004-nltk} essential language processing utilities, while a Cantonese Universal Dependencies dataset \citep{wong-etal-2017-quantitative} offers a small yet significant syntactically annotated treebank.
Multilingual resources such as Wikipedia \citep{wikidump}, SIB-200 \citep{adelani-etal-2024-sib}, and NLLB \citep{nllb-24-flores200} include Cantonese portions, which may be useful for representation learning or evaluation in topic classification and multilingual translation.
Larger-scale corpora such as YueData \citep{jiang2025developing} have also emerged, though they often contain substantial portions of non-Cantonese text.

Beyond corpora, work has extended to specific applications, including sentiment analysis \citep{ZHANG2011OpenRice}, automatic speech recognition \citep{li2019cantonese}, machine translation \citep{liu-2022-low, suen-etal-2024-leveraging, dai-etal-2025-next, liu2025siniticmterror}.
Most recently, \citet{cheng-etal-2025-hkcanto} introduced HKCanto-Eval, a comprehensive benchmark for evaluating linguistic and cultural understanding in Cantonese.
\citet{jiang-etal-2025-well} presents another benchmark for evaluating LLM reasoning, knowledge, and logic in Cantonese.

On the modeling side, prior work has explored both general-purpose and Cantonese-specific pretrained models.
While commercial multilingual models such as Qwen \citep{bai2023qwen} and DeepSeek \citep{liu2024deepseek} provide Cantonese support, their Cantonese proficiency lags behind that in English or Mandarin \citep{jiang-etal-2025-well}.
\citet{jiang2025developing} use their \texttt{YueData} corpus to train \texttt{YueTung-7b}, a continually pre-trained model based on a Qwen-7B model \citep{bai2023qwen}, which exhibit improved Cantonese performance compared to other open-source and commercial LLMs.
On the smaller end in terms of the number of parameters, \citet{cheng2024bertbasecantonese} represents the only encoder-only Cantonese model to our knowledge.
It is continually pre-trained \texttt{bert-base-chinese} on Cantonese news articles, social media posts, and web pages on.
Implementation details, training recipes, and corpora selection for \texttt{bert-base-chinese} and \texttt{bert-base-cantonese} are both obscure as neither model provides a description paper or technical report. In particular, \citet{devlin-etal-2019-bert} describes the BERT architecture in general but not the Chinese model.

\paragraph{\textsc{CantoNLU} and other Cantonese benchmarks.}
Although we have described Cantonese benchmarks HKCanto-Eval \citep{cheng-etal-2025-hkcanto} and Yue-Benchmark \citep{jiang-etal-2025-well} as related work, we wish to clarify our contributions to Cantonese benchmarking in comparison to these two benchmarks.
While HKCanto-Eval includes OpenRice \citep{ZHANG2011OpenRice} as a SA dataset, which overlaps with our \textsc{CantoNLU}, along with minor datasets in Cantonese phonology and orthography, HKCanto-Eval and Yue-Benchmark primarily target generative LLMs on general world knowledge and reasoning, in a comparable way to MMLU \citep{hendrycks2021measuring}.
We highlight \textsc{CantoNLU}'s focus on discriminative NLU tasks, explicitly evaluating a model's ability to understand the Cantonese lexicon (WSD), syntax (POS, DEPS), semantics (NLI, SA), and overall well-formedness with respect to both syntax and semantics (LAJ).
The focus is similar to those of GLUE \citep{wang-etal-2018-glue} and its derivatives--in Korean \citep{park2021klue}, Chinese \citep{xu-etal-2020-clue}, Vietnamese \citep{do-etal-2024-vlue}, and Indonesian \citep{wilie-etal-2020-indonlu}.


\section{Building \textsc{CantoNLU}}
\label{sec:benchmark}
We introduce a benchmark of seven Cantonese NLU tasks, encompassing word sense disambiguation (WSD), linguistic acceptability judgment (LAJ), language detection (LD), natural language inference (NLI), sentiment analysis (SA), part-of-speech tagging (POS), and dependency parsing (DEPS).
While all tasks require Cantonese proficiency, tasks such as LD may reward Mandarin knowledge, whereas tasks such as DEPS and WSD may penalize Mandarin knowledge.
WSD, LAJ, and LD datasets are novel contributions to Cantonese NLU.

\paragraph{Novel WSD dataset via manual compilation.}
First, we manually compile Cantonese words with more than one attested meaning to create the first Cantonese word sense disambiguation dataset.
We collect the words' multiples senses and two example sentences for each sense, resulting in 41 multi-sense words with a total of 109 senses.
For each sense, there are least 2 example sentences containing the word.
The dataset does not require fine-tuning for evaluation—model predictions are obtained by masking the target word in each sentence and comparing cosine similarities between the hidden representations at the mask position.
For each target word $w$ with two contexts $s_i$ and $s_j$, we obtain hidden representations 
for each at the masked position from the model $\mathbf{h}_i$ and $\mathbf{h}_j$.  

Using cosine similarity, we define same-sense score $s_{\text{same}}$ and different-sense score $s_{\text{diff}}$ as :
\[
s_{same} = 
\frac{1}{|P_{\text{same}}|} 
\sum_{(i,j) \in P_{\text{same}}} 
\text{cos}(\mathbf{h}_i, \mathbf{h}_j),
\]
\[
s_{\text{diff}} = 
\frac{1}{|P_{\text{diff}}|} 
\sum_{(i,j) \in P_{\text{diff}}} 
\text{cos}(\mathbf{h}_i, \mathbf{h}_j),
\]
where $P_{\text{same}}$ and $P_{\text{diff}}$ are the sets of sentence pairs containing 
the same and different senses of $w$, respectively. A model prediction is considered correct if $s_{\text{same}} > s_{\text{diff}}$.

\paragraph{Novel LAJ dataset adapted from error span dataset.}
LAJ is a classification task, where the model predicts whether the given sequence is linguistically acceptable.
This often aligns with grammatical judgment acceptability, but may also include judgment on semantic plausibility or pragmatic felicity.
We compile the first Cantonese LAJ dataset by adapting the Cantonese portion of \textsc{SiniticMTError} \citep{liu2025siniticmterror}, a dataset of Sinitic translation error span annotations, where each datapoint consists of a well-formed reference sentence \texttt{ref}, a machine translated sentence \texttt{mt}, and annotations of errors in the \texttt{mt} sentence.
We consider error-free \texttt{ref} as acceptable, \texttt{mt} with error annotations not acceptable, to create pairs with one acceptable and one not acceptable versions of the same sentence.
Unlike previous LAJ datasets such as CoLA \citep{wardtstadt2019cola}, which asks for a binary acceptable-not acceptable judgment, we implement a more robust setup of providing two versions of the same sentence and asking for a more acceptable version as is preferred in psycholinguistics and cognitive science \citep{mahowald2016snap, linzen2018reliability}.
The dataset consists of 1.6k pairs.

\paragraph{Novel LD dataset from word-aligning and perturbing parallel corpus.} 
Language Detection (LD) is a three-label classification task that identifies whether a given sentence is written in Cantonese, Mandarin, or mixed.
We construct the novel dataset from the parallel translation corpus of \citet{dai-etal-2025-next}, selecting the first 10,000 sentence pairs.
To create mixed-language examples, we randomly replace tokens in Cantonese and Mandarin sentences with their counterparts from the other language with a set probability. 
In a given mixed sentence, 15\%, 33\%, 50\% of the sentence may be from the other language, thus producing up to 6 mixed sentences for each pair.
Word-level alignments are obtained using SimAlign \citep{jalili-sabet-etal-2020-simalign}, and all text is converted to traditional orthography using HanziConv \citep{hanziconv} to prevent script-based shallow heuristics.
The resulting dataset contains 47,578 sentences, of which 27,578 are mixed.
We reserve 5\% each for the validation and testing splits. 
We note that this task requires proficiency in both Mandarin and Cantonese to perform well.

\paragraph{Machine translated English NLI datasets.}
NLI is a classification task where the model is asked to predict whether the premise entails, or implies the truth of, the hypothesis. 
The NLI portion of our benchmark is the \texttt{yue-nli-all} dataset \citep{cheng2025yue-all-nli}, which is a machine translation of English NLI datasets SNLI \citep{bowman-etal-2015-large} and MNLI \citep{williams-etal-2018-broad} into Cantonese.
The dataset comprises of  557k train examples, 6.6k development examples, and 6.6k test examples.
Each example includes a reference premise and two hypotheses, one entailing and one contradicting, from which we create two examples with each label.
This results in a balanced, two-label classification dataset.

\paragraph{Sentiment analysis dataset from restaurant reviews.}
Sentiment analysis is a sentence-level classification task of predicting the sentiment of a given sentence.
We use the OpenRice dataset \citep{ZHANG2011OpenRice} compiled from restaurant reviews in Hong Kong, with the 3-way label space of positive, neutral, and negative.
The dataset is balanced, with 10k datapoints in its train split, 1k in its development split, and 1k in its test split.

\paragraph{POS and DEPS from Cantonese UD.}
Finally, POS and DEPS are both token-level classification tasks.
A POS model predicts the part-of-speech tag (e.g. noun, verb, etc.) of a given word, while a DEPS model predicts the dependency head (answering, which word is the syntactic head of this word?) and dependency type (answering, what is the relationship between this word and its head?) of the given word.
For POS and DEPS, we use the Cantonese-HK Universal Dependency dataset \citep{wong-etal-2017-quantitative}, which comprises of 1k sentences and 14k tokens.
We split the dataset 9:1 for training and testing, respectively.
The dataset contains 15 POS tags and 48 dependency relations, of which 17 are specific to Cantonese.
We report both unlabeled (UAS) and labeled attachment scores (LAS) to measure DEPS performance.

\section{Experimental Benchmark Setup}
\label{sec:exp}
Using the compiled Cantonese NLU benchmark, we evaluate three types of models--Cantonese monolingual, Cantonese-adapted from Mandarin, and Mandarin.

\paragraph{Pre-training corpora.}
For those requiring training or adaptation into Cantonese, we use two corpora: the Cantonese Wikipedia \citep{wikidump} and a list of 30 million Cantonese sentences compiled by \citet{kwok-2024-cantonese-sentences}.
The two corpora are the publicly available and open-source subset of YueData \citep{jiang2025developing}.
The Wikipedia dump includes empty parentheses from their pre-processing stage that removes text in other languages.
For example, (5) becomes (6), whose process often yields parentheses that are either empty or only contain punctuation marks.
We remove them.
Then, the corpora are divided into excerpts of maximum length 128 for pre-training.
The Cantonese Wikipedia contains 137k articles, totaling 40M characters, while \citet{kwok-2024-cantonese-sentences} contains 30M sentences, totaling 660M characters.
In total, the pre-training corpora consists of 700M characters.

\begin{CJK*}{UTF8}{gbsn}
\begin{enumerate}
    \item[(5)] 電影（英語：movie/ film），特点是运动／移动的画面（英語：motion/ moving picture)\footnote{The illustrated example is from Mandarin, due to LaTeX compilers' difficulty with Cantonese text.} \\
    \item[(6)] 电影（/ ），特点是运动／移动的画面（）
\end{enumerate}
\end{CJK*}
\paragraph{Cantonese-adapted model.}
The Cantonese-adapted model represents the most common text processing effort in Cantonese, taking an existing model with Mandarin support and performing continued pre-training on Cantonese before applying the adapted text or speech model to downstream tasks \citep{luo2021crosslanguage, cheng2024bertbasecantonese, jiang2025developing}.
In our implementation, we take the off-the-shelf \texttt{bert-base-chinese} model \citep{devlin-etal-2019-bert}\footnote{The citation is attributable to the \texttt{bert} language model as a whole; it does not include implementation details on \texttt{bert-base-chinese}.} and continually pre-train on Cantonese text described above.
We do not make any changes to the model's tokenizer.

In addition, we offer a comparison to a concurrent BERT-based Cantonese work in \texttt{bert-base-cantonese} \citep{cheng2024bertbasecantonese}, also a continually pre-trained model, but on a close-source corpus of news articles, social media posts, and web content.

\paragraph{Mandarin model without Cantonese adaptation.}
Direct transfer from Mandarin without conditioning on Cantonese also represents a small subset of Cantonese NLP efforts \citep{liu-2022-low, li2024finetuning}, including evaluating non-Cantonese trained multi-lingual LLMs such as Llama models on Cantonese knowledge benchmarks \citep{cheng-etal-2025-hkcanto}.
We employ \texttt{bert-base-chinese} \citep{devlin-etal-2019-bert} as a model representing direct transfer from Mandarin, and fine-tune the model on the downstream tasks without additional conditioning on Cantonese text.
We do not make any changes to the model's tokenizer.

\paragraph{Monolingual Cantonese model.}
We train a monolingual Cantonese model from scratch using the BERT architecture \cite{devlin-etal-2019-bert}.
While previous work in Cantonese language modeling have incorporated additional datasets \citep{jiang2025developing}, many of them are proprietary, may include non-Cantonese text, or may not be freely used as pointed out by \citet{xiang-etal-2024-cantonese}.
Thus, similar to the Cantonese-adapted model, we train a monolingual Cantonese model using the publicly available Cantonese Wikipedia dump \citep{wikidump} and the \texttt{cantonese-sentences} corpus \citep{kwok-2024-cantonese-sentences}.
The amount of data, totaling 700M characters, is an order of magnitude smaller than what was used to train \texttt{bert-base-uncased} at 3.3B words \citep{devlin-etal-2019-bert}.
We are unable to make an exact comparison to \texttt{bert-base-chinese} as training details of the model are not publicly available, although we suspect the Mandarin Wikipedia dump was used for a part of its training, which consists of around 1 million articles (cf. 137k articles for Cantonese) at the time of \texttt{bert-base-chinese} training.

For the monolingual model, we train a sentencepiece byte-pair encoding tokenizer on the same data to obtain a Cantonese-only tokenizer.
Unlike the tokenizer from \texttt{bert-base-chinese} which only includes 8k tokens of character length 1, the Cantonese tokenizer captures subword structure by including around 32k tokens, of which 27k are multi-character tokens. Moreover, a high overlap in lexicons is maintained, with 4.8k characters being represented in both \texttt{bert-base-chinese} and our Cantonese tokenizer.
While we expected greater coverage of Cantonese-only characters from Hong Kong Supplementary Character Set (HKSCS), this is not the case as \texttt{bert-base-chinese}'s tokenizer boasts a greater coverage with 603 characters, while our tokenizer covers only 233 characters in HKSCS.
This result reflects the relative scarcity of Cantonese-specific data compared to the abundant Mandarin data available.

Each model is fine-tuned on the training split of the downstream NLU task, then evaluated on the test split of the same task with the exception of LAJ and WSD which use model surprisal without fine-tuning.
We report accuracy metrics for NLI, LAJ, WSD; F1 metrics for POS, LD and SA; and UAS and LAS for DEPS.
Following NLU convention \citep{devlin-etal-2019-bert}, we do not freeze model weights and allow them to be updated during fine-tuning.
We report task-specific hyperparameter choice during fine-tuning in Table \ref{tab:hyperparams}.

\begin{table}[th]
\centering
\begin{tabular}{lccc}
\toprule
\textbf{Task} & \textbf{LR} & \textbf{Batch} & \textbf{Epochs}\\
\midrule
NLI                & 2e-5 & 16 & 3 \\
POS        &  2e-5 & 32 & 3 \\
DEP & 3e-5 & 16 & 20 \\
LD & 2e-6 & 16 & 3 \\
SA & 2e-5 & 16 & 3 \\
LAJ & \multicolumn{3}{c}{No fine-tuning performed} \\
WSD & \multicolumn{3}{c}{No fine-tuning performed} \\

\bottomrule
\end{tabular}
\caption{Hyperparameters used for fine-tuning across tasks. LR = learning rate. Patience refers to the number of epochs before early stopping.}
\label{tab:hyperparams}
\end{table}

\begin{table*}[ht]
    \small
    \centering
    \begin{tabular}{r|cccccccc|c}
         \toprule
          & WSD & LAJ & LD & NLI & SA & POS & \multicolumn{2}{c}{DEPS} & Avg. \\
         Model & Acc. & F1 & F1 & Acc. & F1 & F1 & UAS & LAS & \\
         \midrule
         \textbf{No Cantonese adaptation} \\
         \texttt{bert-base-chinese} & 78.9 & \textbf{91.7} & 76.4 & \textbf{93.2} & 70.2 & 74.6 & 29.1 & 25.9 & 67.5\\
         \midrule
         \textbf{Cantonese-adapted} \\
         \texttt{bert-base-cantonese} & \textbf{92.7} & 89.2 & \textbf{78.7} & \textbf{93.2} & \textbf{71.9} & 72.2 & 30.2 & 26.8 & \textbf{69.4} \\
         Our transfer model & 85.3 & 89.6 & 78.4 & 87.5 & 71.3 & 74.8 & 30.0 & 26.8 & 68.0 \\
         \midrule
         \textbf{Monolingual Cantonese} \\
         Our monolingual model & 70.6 & 85.7 & 73.3 & 82.6 & 70.1 & \textbf{78.2} & \textbf{32.4} & \textbf{27.9} & 65.1\\
         \bottomrule
    \end{tabular}
    \caption{Performance of Mandarin, Cantonese-adapted, and monolingual models across Cantonese NLU tasks. We offer performances of open-weight but closed-source models \texttt{bert-base-chinese} \citep{devlin-etal-2019-bert} and \texttt{bert-base-cantonese} \citet{cheng2024bertbasecantonese} as comparison.}
    \label{tab:main-results}
\end{table*}

\section{Results and Discussion}
\label{sec:results}

Our results across the 7 Cantonese NLU tasks are shown in Table~\ref{tab:main-results}, where our Cantonese monolingual model demonstrates the highest performance for POS and DEPS; Cantonese-adapted for NLI, LD, and WSD; Mandarin model without Cantonese adaptation for NLI and LAJ.
Our transfer model's performances slightly tail that of \texttt{bert-base-cantonese}.
We highlight three main findings. 
First, Cantonese monolingual models excel in syntactic tasks, while Cantonese-adapted Mandarin models are currently the most effective approach for Cantonese semantic tasks.
Second, Mandarin-only models can still perform competitively on some tasks.
Finally, despite the relative success of monolingual and transfer models, there remains substantial room for improvement in representing Cantonese lexicon, syntax, and semantics.

\paragraph{Transfer from Mandarin is the most effective for Cantonese NLU.}
Supporting the empirical success of Mandarin-to-Cantonese transfer seen in contemporary Cantonese NLP, Cantonese-adapted Mandarin models offer the strongest performance with a task-averaged score of 69.4 (\texttt{bert-base-cantonese}) and 68.0 (our open-source transfer model).
While the monolingual model excels in syntactic tasks POS and DEPS, its average score of 65.1 lags behind those the Cantonese-adapted and Mandarin models.
We attribute this primarily to the small size of our Cantonese pretraining corpus (roughly 700M characters).
As discussed in Section~\ref{sec:exp}, this is an order of magnitude smaller than the datasets used to train comparable English or Mandarin models, and therefore insufficient for learning robust linguistic representations from scratch.

\paragraph{A well-trained Mandarin model may be sufficient for some Cantonese NLP}
Interestingly, \texttt{bert-base-chinese} achieves comparable, or in some cases, superior performance to its Cantonese-adapted and Cantonese-monolingual counterparts with an average score of 68.1.
When scholars discuss mutual intelligibility among Sinitic languages, they typically refer to the spoken form \citep{tang2007mutual, gooskens2021mutual}.
However, written Sinitic languages are far more mutually intelligible, given the historical influence of Mandarin as the dominant language of education and literacy across Chinese-speaking regions \citep{snow2004cantonese, xiang-etal-2024-cantonese}.
This suggests that effective written Cantonese understanding can emerge from training on Mandarin text without explicit exposure to Cantonese text, as also observed in \citet{jiang2025developing}, where LLMs without explicit Cantonese support perform relatively well on Cantonese and Hong Kong-related knowledge benchmarks.

\paragraph{Despite effective transfer, Cantonese representations remain limited.}
While models achieve respectable performance on NLI, POS tagging, and lexical disambiguation, dependency parsing remains notably weak—likely due to both the small fine-tuning dataset of around 1k sentences and the inherent difficulty of the task.
Nonetheless, dependency parsing for other low-resource languages such as Buryat \citep{badmaeva2017buryat} and Old English \citep{levine-etal-2025-building} has reached higher performance with smaller datasets, suggesting that current Cantonese representations still lack sufficient syntactic and semantic grounding.

Taken together, these results point to the promise of continued pre-training for Cantonese NLP and to the need for richer, higher-quality Cantonese corpora to close the representational gap.

\section{Limitations}
While we proposes a novel evaluation framework and recommend insights into Cantonese representation learning, several limitations remain.
First, the size and coverage of available Cantonese corpora significantly constrain our results.
Our monolingual Cantonese model was trained on roughly 700M characters, an order of magnitude smaller than corpora typically used to pre-train large language models in other major languages \citep{devlin-etal-2019-bert}.
This data sparsity likely limits the model's ability to capture the full lexical and semantic diversity of written Cantonese, and may explain the comparatively weaker performance of the monolingual model in semantic tasks. Future work would benefit from expanding and diversifying Cantonese text resources, especially in informal and user-generated domains that reflect contemporary usage.

Second, our evaluation benchmark, though designed to cover multiple aspects of Cantonese NLU, is itself constrained by data availability. Some tasks, such as dependency parsing (DEPS), rely on small fine-tuning datasets \citep{wong-etal-2017-quantitative}, making the results more susceptible to statistical noise and limiting generalization. In addition, the benchmark focuses on written Cantonese and does not address spoken or colloquial aspects of the language such as code-switching \citep{YIM_BIALYSTOK_2012} or informal register in-depth; these aspects are integral to Cantonese as it is a primarily spoken language \citep{snow2004cantonese}.

As a result, while our findings suggest that Mandarin-to-Cantonese transfer is effective for semantic tasks and Mandarin models perform sufficiently well, they should be interpreted as a reflection of current data and resource disparities, rather than representative features of Mandarin and Cantonese.

\section{Conclusion}
\label{sec:conclusion}
In this paper, we describe a benchmark of Cantonese NLU tasks, and evaluate a monolingual Cantonese model, two transfer models from Mandarin, and a Mandarin model on said Cantonese benchmark to investigate contexts where each type of model is most effective.
Our results indicate that both Cantonese monolingual models and Cantonese-adapted models with cross-lingual transfer from Mandarin both have merit for Cantonese NLP in today's landscape of language resources.
In addition, direct transfer from a Mandarin model without Cantonese representation learning may suffice for some tasks. 
Our findings also suggest that the existing open-source Cantonese corpora are insufficient to train a reliable representation of Cantonese lexicon, syntax, and semantics.

Our benchmark and analyses provide the first systematic investigation into and evidence of whether and how transfer from Mandarin is effective for performing Cantonese linguistic tasks.
By establishing a framework and  pipeline for training monolingual or transfer models and evaluating them, we hope to catalyze broader progress in the space of Cantonese NLP.

Transfer may light the way, but data will pave the road.



\section*{Responsible Research Statement}
While we do not collect human annotations for our work, we acknowledge that we make use of datasets that were annotated by humans or otherwise adapted from a human source.

We make use of a variety of Cantonese NLP and language resources.
In addition to our discussion of them in Sections \ref{sec:related-work}, \ref{sec:benchmark}, and \ref{sec:exp}, 
we acknowledge their use and organize them below.
\begin{itemize}
    \item \texttt{bert-base-chinese}  \url{huggingface.co/google-bert/bert-base-chinese}
    \citep{devlin-etal-2019-bert}
    
    \item \texttt{Cantonese Wikipedia} \url{huggingface.co/datasets/wikimedia/wikipedia/viewer/20231101.zh-yue}
    \citep{wikidump}
    
    \item\texttt{Cantonese-HK UD Corpus}
    \url{github.com/UniversalDependencies/UD_Cantonese-HK}
     \citep{wong-etal-2017-quantitative} 
     
    \item\texttt{Cantonese Sentences} \url{huggingface.co/datasets/raptorkwok/cantonese_sentences}
    \citep{kwok-2024-cantonese-sentences} 
    
    \item\texttt{Yue-All-NLI}
    \url{huggingface.co/datasets/hon9kon9ize/yue-all-nli}
      \citep{cheng2025yue-all-nli} 

    \item\textsc{SiniticMTError}
    \citep{liu2025siniticmterror}

    \item OpenRice
    \url{/www.openrice.com/en/hongkong}, collected by \citet{ZHANG2011OpenRice}

    \item Parallel corpus from
    \citet{dai-etal-2025-next}
      
    \item\texttt{cantonese-chinese-parallel-corpus}
    \url{huggingface.co/datasets/HKAllen/cantonese-chinese-parallel-corpus}

    \item\texttt{Hong Kong Cantonese Corpus}
    \url{github.com/fcbond/hkcancor}
    \citep{luke2015hong}

    \item\texttt{Words.HK}
    \url{words.hk}
    \citep{lau2022words}

    \item\texttt{CC-Canto}
    \url{cantonese.org}

    \item\texttt{CEDict}
    \url{cedict.org}
    
    \item\texttt{KaiFang CiDian}
    \url{kaifangcidian.com}

    \item\texttt{Tatoeba}
    \url{tatoeba.org}

    \item Hong Kong Supplementary Character Set (HKLSCS) \url{www.ccli.gov.hk/en/hkscs/what_is_hkscs.html}
\end{itemize}

Finally, we disclose that we use ChatGPT\footnote{\url{https://chat.openai.com}} and PyCharm's AI\footnote{\url{https://www.jetbrains.com/pycharm/features/ai/}} as a writing and coding assistant during the project's implementation and the paper's writeup.


\section*{Bibliographical References}
\bibliographystyle{lrec2026-natbib}
\bibliography{custom}


\appendix

\end{document}